\documentclass[runningheads]{llncs}
\usepackage{graphicx}
\usepackage[utf8x]{inputenc}
\usepackage[english,turkish]{babel}
\usepackage[numbers]{natbib}
\newcommand*\samethanks[1][\value{footnote}]{\footnotemark[#1]}

\newenvironment{keywords}
{\par\noindent\small\textbf{\keywordsname:}}
{\par}
\addto\captionsenglish{\def\keywordsname{Keywords}}
\addto\captionsturkish{\def\keywordsname{Anahtar kelimeler}}

\makeatletter
\def\blfootnote{\xdef\@thefnmark{}\@footnotetext}
\makeatother

\usepackage{amssymb}

\begin{document}
\title{Koşullu Çekişmeli ve Evrişimli Sinir Ağ Tabanlı Otomatik Karaciğer Bölütleme}
\author{Bora Baydar\inst{1}\thanks{İki yazar da eşit yenilik sunmaktadır.} \and
	   Savaş Özkan\inst{1}\samethanks[1] \and
    	   Gözde Bozdağı Akar\inst{1}}
\authorrunning{B. Baydar et al.}
\institute{Elektrik/Elektronik Mühendisliği, Ortadoğu Teknik Üniversitesi, 06800, Ankara, Türkiye }
          %\email{borisbaydar@gmail.com, \{ozkan.savas, bozdagi\}@metu.edu.tr} }

\blfootnote{Email: \url{ozkan.savas@metu.edu.tr}}

\maketitle              % typeset the header of the contribution
\selectlanguage{turkish} 
\begin{abstract}
Tıbbi görüntülerin otomatik bölütlenmesi, bu alandaki uzmanlara zaman kazandırması ve insan hata faktörünü azaltması sebebiyle tıp bilişimi alanında revaçta olan çalışmalardan biridir. Bu bildiride karaciğer MR görüntülerinin otomatik bölütlemesi için Koşullu Çekişmeli Ağ (KÇA) ve Evrişimli Sinir Ağ (ESA) tabanlı otomatik yeni bir yöntem sunulmaktadır. Önerilen yöntem, hiçbir son işleme olmadan, Dokuz Eylül Üniversitesi tarafından sağlanan veriler üzerinden düzenlenen SIU Karaciğer Bölütleme Yarışması 2018'de en başarılı 2. sonucu elde etmiştir. Bu bildiride, son işleme adımında yeni iyileştirmeler önerilmiş ve önerilen eklemelerle, yöntemin diğer yöntemlere kıyasla daha başarılı sonuçlar üretebildiği gösterilmiştir. 	
	
%Tıbbi görüntülerin otomatik bölütlenmesi, bu alandaki uzmanlara zaman kazandırması ve insan hata faktörünü azaltması sebebiyle tıp bilişimi alanında en revaçta olan çalışmalardan biridir. Bu çalışmada, Dokuz Eylül Üniversitesi tarafından sağlanan karaciğer MR görüntülerinin otomatik  bölütlemesi için Koşullu Çekişmeli Sinir Ağı (KÇA) ve Evrişimli Sinir Ağ (ESA) tabanlı otomatik yeni bir yöntem sunulmaktadır. Önerilen yöntem, hiçbir son işleme olmadan, SIU Karaciğer Bölütle-me Yarışması 2018'de en başarılı 2. sonucu elde etmiştir. Fakat, son işleme adımında sunulan eklemelerle, önerilen yöntemin diğer yöntemlere kıyasla oldukça başarılı sonuçlar üretebildiği sunulmaktadır.

\end{abstract}
%\selectlanguage{turkish}
%\keywords{Karaciğer Bölütleme \and Tam Evrişimsel Sinir Ağları \and Koşullu Çekişmeli Sinir Ağlar}
\selectlanguage{english} 
\begin{abstract}
Automatic segmentation of medical images is among most demanded works in the medical information field since it saves time of the experts in the field and avoids human error factors. In this work, a method based on Conditional Adversarial Networks and Fully Convolutional Networks is proposed for the automatic segmentation of the liver MRIs. The proposed method, without any post-processing, is achieved the second place in the SIU Liver Segmentation Challenge 2018, data of which is provided by Dokuz Eylül University. In this paper, some improvements for the post-processing step are also proposed and it is shown that with these additions, the method outperforms other baseline methods. 
\end{abstract}

\section{Giriş}

Son yıllarda, tıbbi görüntü analizi süresince uzmanlarca harcanan zamanı ve insan hatalarını azaltmak için otomatik bilgisayar tabanlı yaklaşımlar sıklıkla tercih edilmektedir. Otomatik nesne bölütleme özellikle birçok tıbbi görüntüleme arasında en önemli ve ihtiyaç duyulan uygulama alanı olarak belirlenebilir. Kısaca bölütleme, görüntünün/sahnenin anlamlı bütünlük sağlayacak şekilde parçalara ayrılması işlemidir ve hastalıkların saptanmasında, tedavinin planlamasında ve tedavi sonrası gelişimi izlemede yaygın olarak kullanılmaktadır.

MR görüntülerinin bölütlemesi için literatürde birçok farklı yöntem sunulmuş-tur  \cite{hyperDenseNet},\cite{DeepMedic/KamnitsasLNSKMR16},\cite{BraTS2017_135_3DU_Net},\cite{BraTS2017_100_3DU_Net},\cite{BraTS2017_297_anisotropic}
. Özellikle, veri miktarı ve paralel işleme gücündeki artış, derin öğrenme tabanlı yöntemlerin daha etkin bir şekilde kullanılabilmesini hızlandır-mıştır. Evrişimli sinir ağ yapısı, sahne sınıflandırma için etkin olarak kullanılmak-ta \cite{Krizhevsky},\cite{ResNetDBLP:journals/corr/HeZRS15},\cite{googleNet43022},\cite{VGG_DBLP:journals/corr/SimonyanZ14aVGG16} ve yalnızca son ağ katmanında yapılan değişikliklerle bölütleme problemine uyarlanabilmektedir \cite{fullyConvNetsSemanticSeg}. Literatürde, bu konuda çok başarılı yakla-şımlar önerilmiştir. Dolayısıyla en son sunulan tıbbi görüntü bölütleme yöntemlerinin büyük bir kısmı Tam Evrişimli Ağ(TEA) tabanlıdır. 

Bu çalışmada, MR görüntülerinin yüksek başarım oranıyla otomatik olarak bölütlenmesi için yenilikçi bir yöntem önerilmektedir. Bu kapsamda, çalışma uzayı olarak MR karaciğer görüntüleri yapılan testlerde kullanılmıştır. Önerilen yöntemde, tam evrişimli ağ yapısı tercih edilmiştir. Özellikle, literatürdeki genel filtre yapısından farklı olarak önemli yapısal değişiklikler sunulmaktadır. Bu sayede, ağ parametrelerinin aşırı öğrenme oranları engellenmiştir. Ayrıca, çekişmeli sinir ağ öğrenme adımı \cite{Generative_Adversarial_NIPS2014_5423}, sınıflandırma adımı ile birlikte kullanılmış, böylelikle evrişimsel ağların MR nesne sınırlarında yumuşak olan bölütleme so-nuçları daha keskin olması sağlanmıştır. Önerilen yöntem, MR kanallarını tek ya da küçük kanal grupları olarak işleyebildiğinden, bölütme sonuçlarında oluşabile-cek ayrışımları gidermek için son filtreleme adımı uygulanmıştır. Bu sayede, elde edilen MR bölüt görüntüsü tek bir bütün nesne parçası olarak hesaplamaktadır.

Bildirinin devamında, ilk olarak literatürde MR bölütme üzerine yapılan araştırmalar özetlenmektedir. Daha sonra, bildiride önerilen yöntem detaylı olarak anlatılmaktadır. Son olarak, deneyler sırasında kullanılan veriseti, yapılan testler ve elde edilen sonuçlar karşılaştırmalı olarak sunulmaktadır.

\section{İlgili Çalışmalar}
Ronneberger \textit{et al.}, \cite{fullyConvNetsSemanticSeg} de sunulan tam evrişimli ağı (FCN-8s) geliştirerek tıbbi görüntü bölütlemesi için U-Net'i sunmuştur \cite{U_Net}. U-Net, atlama bağlantıları ve adım adım üst örnekleme yapması sayesinde performansı artırmış ve sunulduğu dönemden itibaren, tıbbi görüntü bölütlemede araştırmacıların tercihi haline gelmiştir. Buna rağmen, doğru değişiklikler ve ağ yapısının probleme uyarlanması ile hala geliştirmeye açıktır. 

Beyin MR bölütlemesi gibi çok sınıflı problemler için literatürde birçok yöntem bulunmaktadır. Bu yöntemler bazen farklı ağlar kullanarak performansı artırmayı hedeflerken \cite{BraTS2017_297_anisotropic,BraTS2017_135_3DU_Net}, bazıları ise ESA'nın iç yapısını değiştirerek bunu başarmaya çalışmıştır \cite{BraTS2017_154_inception,GAN_DilatedConv,BraTS2017_175_3DU_Net}. Örneğin, \cite{BraTS2017_297_anisotropic} kademeli bir ağ yapısı önerilmiştir. Bu yapıda herbir sınıfın sınıflandırması sıralı bir şekilde yapılmaktadır. Buna karşın \cite{BraTS2017_135_3DU_Net} farklı yapıda birçok ağı eğitip bunların sonuçlarını birleştirerek son sınıf etiketine ulaşmıştır. 

ESA'ların performansını artırmak için önceden sunulan bazı yapısal değişiklik-ler MR bölütlemesine de uygulanmıştır. Bunlardan öne çıkanlar başlangıç modül-leri \cite{googleNet43022}, genişletilmiş evrişimler \cite{contextModuleDBLP:journals/corr/YuK15} ve artık bağlantı blokları \cite{ResNetDBLP:journals/corr/HeZRS15} olarak özetlene-bilir. Bunlara ek olarak, birçok problemde olduğu gibi MR bölütleme probleminde de aktivasyon fonksiyonu, kayıp fonksiyonu vb. seçimlerin performansı önemli ölçüde etkileyebileceği bilinmektedir.

\section{Tıbbi Görüntü Bölütleme için Genelleştirilmiş Bir Derin Öğrenme Yaklaşımı}
%{A Generalized Deep Learning Approach for Medical Image Segmentation}

Bu çalışmadaki hedefimiz, bütün MR görüntü bölütleme problemlerinde iyi sonuç verebilecek, genelleştirilmiş bir yaklaşım sunmaktır. Ana bölütleme ağı olarak, FCN-8'e göre avantajları olduğu için U-Net kullanılmıştır. Bu avantajlardan en önemlisi U-Net'in atlama bağlantılarını kullanarak hatanın daha kolay yayılım yapmasını sağlamasıdır. Bu sayede bilgi kaybetme problemi azaltılmış olur. 

\begin{figure*}[t]
	\centering
	{\shorthandoff=%
		\includegraphics[scale=0.3]{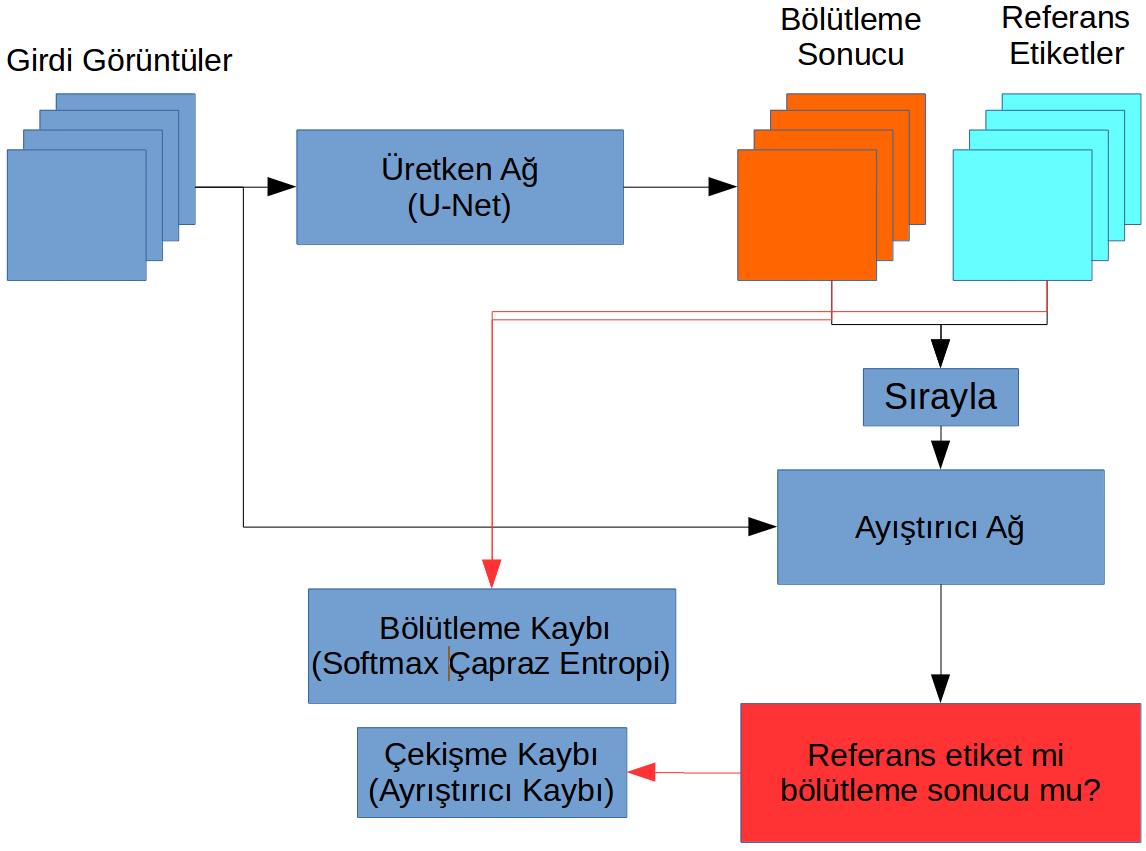}}
	\caption{Önerilen yöntemin genel akışı \label{Fig:generalNetwork}}
\end{figure*}

Ayrıca, bölütleme ağının eğitimini iyileştirmek için, bu ağ yapısına ek olarak bir çekişmeli ağ kayıp değeri eklenmiştir. Genel ağ yapısı Şekil \ref{Fig:generalNetwork}'de verilmiştir. 

\subsection{Veri Normalleştirme}
Tıbbi veriler toplanırken, birçok farklı kuruluş, hastane, enstitü vs. bu işleme katkı sağlayabilir. Benzer şekilde, bu kuruluşlar farklı MR makineleri (çözünürlük ve kalite farklılıkları) ve farklı protokoller ile bu verileri toplayabilir. Önerilecek yöntem, aynı tipte fakat farklı karakteristiklerdeki verileri aynı başarı oranıyla işleyebilmelidir. Bu nedenle, görüntülerin bir ön işleme adımıyla düzeltilmesi elde edilecek başarıya katkı sağlamaktadır. 

Önerilen ön işleme adımında, her bir MR görüntü, piksel değerleri $x_i \in \rm I\!R^{W,H,D}$  dağılımına göre normalize edilmiştir. Burada, $W$ ve $H$ görüntünün uzamsal boyutlarını gösterirken, $D$ MR görüntünün kanal sayısına karşılık gelmektedir. Bu amaçla, görüntüdeki en yüksek ve en küçük pixel değeri ile görüntü tekrar ölçeklenmiştir:
\begin{eqnarray}
\label{eqn:normalizasyon}
\hat x_i = \frac{x_i - min_x}{max_x - min_x}
\end{eqnarray}

\noindent Burada $max_x$ ve $min_x$ değerleri görüntüdeki bütün pikseller arasında en büyük ve en küçük değerlere karşılık gelmektedir. Bu adım sayesinde, farklı piksel aralıklarındaki görüntüler, aynı değer aralığına ölçütlendiği için bölütlemedeki doğru sınıflandırma oranı artmaktadır.  

\subsection{Genel Evrişimsel Ağ Yapısı}

Bölütleme problemini, sahne sınıflandırma probleminden ayıran en temel fark, çıktı sonucunun tek bir değer yerine herbir piksel için farklı  sonuçlara karşılık gelmesidir. Bu nedenle de, çıktı sonucu girdi görüntüyle aynı uzaysal boyutlara sahip olmalıdır ve bölütleme problemi için her pikselin sınıflandırması yapılma-lıdır. Sınıflandırma problemlerinde, derin katmanlara indikçe uzaysal boyutlar küçülürken, saklı (latent) öznitelik bilgi miktarı artmaktadır. Bölütleme için önerilmiş olan FCN-8 ve U-Net gibi ağlarda \cite{fullyConvNetsSemanticSeg,U_Net}, derin katmanlarda elde edilen ve uzaysal boyutları küçük olan öznitelikler, farklı yollarla üst örnekleme yapılarak girdi boyutlarına yükseltilir. FCN-8 bu işlemi tek adımda yaparken, U-Net bu işlemi herbir katmanda tekrarlı şekilde yapmaktadır. Eğer, görüntülerden öznitelik çıkarma adımına kodlama ve bölütlemenin oluşturulması için örnekleme yapılan adıma da çözümleme denirse, kodlama adımının farklı katmanlarında elde edilen değerler, çözümleme adımında herbir katmanda ayrı ayrı birleştirilmektedir (toplama veya bitiştirme yöntemiyle). Bu sayede, girdi görüntüden çıktı sonucuna kadar olan bilginin miktarını maksimuma çıkarmak amaçlanmaktadır ve eğitim sırasında sonuç değerinde oluşabilecek hatanın daha kolay geri yayılım yapabileceği bir kısayol eklenmiş olur. Ayrıca, FCN-8 modeline göre hesaplama yükü katmanlar halinde birleştirme yapıldığı için nispeten azalmıştır. 

Önerilen yöntemin eğitimi sırasında softmax çapraz entropi \cite{DeepLearningBookGoodfellow-et-al-2016} kayıp değeri kullanılmaktadır:
\begin{eqnarray}
\label{eqn:softmax_cross_entropy}
\mathcal{L}_{cls} = - \frac{1}{N} \sum_{i=1}^{N} ( y_i^R  \log y_i^P )
\end{eqnarray}

\noindent Burada $y_i^R$ gerçek bölüt görüntüsünü gösterirken, $y_i^P$ önerilen yöntem tarafından tahmin edilen bölütleme sonucuna karşılık gelmektedir. $N$ ise görüntüdeki piksel sayısını temsil eder.

\subsection{Yapısal Önermeler}

Literatürde, özellikle geri yayılım bilgisinin kaybolmasını engellemek ve ReLU aktivasyon fonksiyonu \cite{ReLUHahnloser2000DigitalSA} ile kullanıldığında seçiciliği arttırmak için Grup (batch) Normalleştirme \cite{batch_normalization} yaygın olarak kullanılmaktadır. Yaygın olarak her katmanda önce  Evrişimsel Filtre, sonrasında Grup Normalleştirme, ve en son ReLU uygulaması tercih edilmektedir. 

Fakat Grup normalleştirme adımının, evrişimsel filtre yapıdan önce uygulanmasının daha başarılı sonuçlar verdiği deneysel sonuçlarda gözlenmiştir. Ayrıca, ReLU aktivasyon fonksiyonu yerine negatif değerlere kısmi izin veren Parametrik ReLU \cite{pReLU_paper} aktivasyon fonksiyonu tercih edilmiştir. Sonuçların iyileşmesindeki temel etken, filtreleme sonucunda elde edilen negatif değerlerin bilgi akışı sırasın-da tamamen kaybedilmemesi ve olabildiğince en yüksek seviyede bütün değer aralıklarının istatistiki olarak değerlendirmeye alınmış olmasıdır.

\subsection{Çekişmeli Sinir Ağları}

Evrişimli filtre yapısındaki en önemli eksik, nesnenin kenar bölgelerinde yumuşak ve belirgin bir şekilde bozuk şekillerin oluşmasıdır. Bu durumu iyileştirmek için literatürde en fazla tercih edilen yöntemlerin başında Koşullu Rastgele Alanlar gelmektedir. Bu fonksiyon pikselin rengine, komşu piksellerin sınıflarına, pikselin pozisyonuna bağlı olarak ilişkiler kurarak bölütme sonuçlarını iyileştirmeyi amaçlamaktadır. Fakat, öğrenme sırasında, evrişimler ağlardaki parametre değer-lerine etki edemediği için, genel yapıdan bağımsız bir ikinci adım olarak tercih edilmektedir. Bu nedenle bizim yaklaşımımızda Koşullu Rastgele Alanlar tabanlı bir yaklaşım yerine, derin ağ modeliyle etkileşime geçebilecek Koşullu Çekişmeli Ağ yaklaşımı entegre edilmiştir.

Kısaca, klasik Üretken Çekişmeli Ağlarda \cite{Generative_Adversarial_NIPS2014_5423}, ağ rastgele Gaussian vektörleri ile beslenir ve ayrıştırıcı-iyileştirici öğrenme modeli sonucunda eğitim veri setinde bulunan görsellere benzeyen yeni sonuçlar elde edilir. Koşullu Çekişmeli Ağlarda ise ağa aynı zamanda belli koşullar girdi olarak verilebilir. Örneğin, \cite{ConditionalAdvGANDBLP:journals/corr/IsolaZZE16} çalışmala-rında, elde edilmek istenen yüz ifadesi girdi olarak ağa sağlanır. Anlamsal bölütle-me probleminde de yaklaşım buna oldukça benzerdir. MR görüntüsü $(x_i)$ ile elle etiketlenen gerçek bölüt alanı $(y_i^r)$ bitiştirilmiştir ($u_i^R = [x_i, y_i^r]$ , $[.]$ bitiştirme işlemini belirtmektedir). Benzer şekilde MR görüntüsü $(x_i)$ ile üretici ağ tarafın-dan elde edilen sonuç görüntüsü $(y_i^r)$ bitiştirilmiştir $(u_i^P = [x_i, y_i^p])$. Koşullamada, üretici ağ, $u_i^P$ ile $u_i^R$ yi ayırt etmeye çalışan bir ayrıştırıcı ağ ile eğitilmektedir. Böylece, eğitim sırasında nesnelerin kenar sınırlarında oluşacak hatanın modele yayılımı ek bir kayıp değeri ile sağlanır:
\begin{eqnarray}
\label{eqn:lossd}
\mathcal{L}_{gan}^d =   \frac{1}{N} \sum_{i=1}^{N}( \log D(u_i^R) ) + \frac{1}{N} \sum_{i=1}^{N} ( \log( 1.0 - D(u_i^P) ) ),
\end{eqnarray}

\begin{eqnarray}
\label{eqn:lossg}
\mathcal{L}_{gan}^g =  \frac{1}{N} \sum_{i=1}^{N} ( \log D(u_i^P)  )
\end{eqnarray}

\noindent Burada, $\mathcal{L}_{gan}^d$ ayrıştırıcı kayıp değeri, gerçek ile elde edilen bölütleme değerini birbirinden olabildiğince ayırt etmeye çalışırken, $\mathcal{L}_{gan}^g$ evrişimsel filtreleri bölütleri olabildiğince gerçek değerlere benzetmeye yani sınır kenar noktalarında keskinleş-tirmeye çalışmaktadır. Not edilmelidir ki, karaciğer bölütlemede kullanılan KÇA, son işleme adımından ziyade, kayıp fonksiyonunun geliştirilmesi olarak düşünüle-bilir.

\subsection{Öğrenme Modeli}

Üst kısımda özetlendiği üzere, parametre öğrenmede iki kayıp değerlerinden faydalanılmaktadır:
\begin{eqnarray}
\label{eqn:combined_loss}
\mathcal{L} = \mathcal{L}_{cls} + 0.01 \times \mathcal{L}_{gan}^g
\end{eqnarray}

\noindent Dikkat edileceği üzere Koşullu Çekişmeli kayıp değeri oldukça düşük tutulmuştur. Bunun nedeni, \cite{towardsPrincipledMethodsforGAN}'de özetlendiği gibi çekişmeli ağlar parametre uzayını ayırırken devamlı olmadığı ve yerel minimumlara takılabileceğinden dolayıdır. Ayrıca, parametrelerin öğrenilmesi için Adam çözümleyicisi \cite{Adam} kullanılmıştır. Öğrenme katsayısı ve toplam iterasyon sayısı 0.001 ve 100K olarak belirlenmiştir. Kodlanması sırasında Tensorflow kod kütüphanesinden yararlanılmıştır.

\subsection{3-Boyutlu Ayrık Giderme}

Önerilen sistem, 3-boyutlu MR görüntüsü bölütleme esnasında, bütün girdi görün-tülerini kullanmak yerine, sıralı gruplar halinde bölüt tahminleri yapmaktadır. Matematiksel olarak, $f(.)$ önerilen evrişimsel filtre modelini belirtirse, elde edilen herbir bölüt sonucu $2\times K +1$ komşuluğundaki görüntüler kullanılarak $y_i^P = f ([\hat x_{i-K}, \hat x_{i-K+1}, ..., \hat x_{i+K-1}, \hat x_{i+K}])$ ile gösterilebilir. Bu nedenle, tahminler sıra-sında bazı 3-boyutlu nesneden kopuk ayrışımlar oluşacaktır. 

Son adım olarak, elde edilen 3-boyutlu MR görüntüleri birbirine bağlı voksellerden oluşacağı bilgisi de göz önünde bulundurularak, küçük yanlış ayrışımları gidermek için en büyük 3 boyutlu bağlı hacim filtrelemesinin performansı artırdığı gözlenmiştir.

\section{Deneysel Sonuçlar}

İlk olarak kullanılan veriseti ve performans hesaplama metrikleri özetlenecektir. Daha sonra, yapılan deneyler ve elde edilen sonuçlar karşılaştırmalı olarak değer-lendirilecektir.

\subsection{Veriseti ve Metrikler}

SIU Karaciğer Bölütleme Yarışması 2018\footnote{\url{https://eee.deu.edu.tr/moodle/mod/page/view.php?id=7873}}
'de de kullanılan BT veri seti 20 hastadan oluşmaktadır\cite{SELVER_1,SELVER_2}. Bunların 6 tanesi eğitim için referans etiketleri ile birlikte sağlanmıştır. Kalan 14 hastanın ise yalnızca tarama görüntüleri verilmiştir. Bölüm \ref{sonuclar}'de yarışmada elde edilen sonuçlar verilmiştir. Verisetinde herbir hasta farklı karaciğer MR kesiti görüntüsünden oluşabilmektedir. 

\begin{table}[!h]
	\centering
	\caption{Son işleme adımı eklenmeden, farklı komşu katman sayıları kullanılarak elde edilen sonuçlar.} \label{Table:NeighborSlicesEffect}
	\vspace{-13pt}
	{\shorthandoff=%
		\includegraphics[width=\columnwidth]{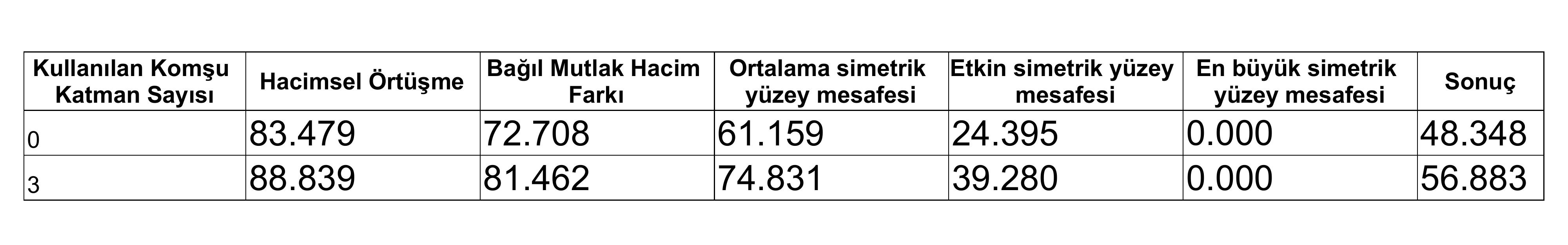}}
	\vspace{-23pt}
\end{table} 

Performans değerlendirmesinde 5 farklı metriğin ortalama değeri dikkate alınmıştır. Bunlar, hacimsel örtüşme, bağıl mutlak hacim farkı, ortalama simetrik yüzey mesafesi, etkin simetrik yüzey mesafesi ve en büyük simetrik yüzey mesafesidir.

\subsection{Sonuçlar} \label{sonuclar}
Önerilen yöntemin sonuçları bu bölümde detaylı olarak sunulmuştur. İlk olarak Tablo \ref{Table:NeighborSlicesEffect}'de son işleme adımı (3B ayrık giderme) olmadan yöntemin sonuçları göstermektedir. Tablodaki kullanılan komşu katman sayısı $K$ değerine karşılık gelmektedir. Sonuçlardan, bölütleme işleminde kullanılan komşu katman sayısının artması sonuçlara olumlu etki yapabildiğini göstermektedir. Fakat, bu artış, hesaplama da benzer bir artışa neden olmaktadır.

\begin{table}[!h]
	\centering
	\caption{2 boyutlu ayrık giderme sonrası elde edilen sonuçlar.} \label{Table:postProcEffect}
	{\shorthandoff=%
		\includegraphics[width=\columnwidth]{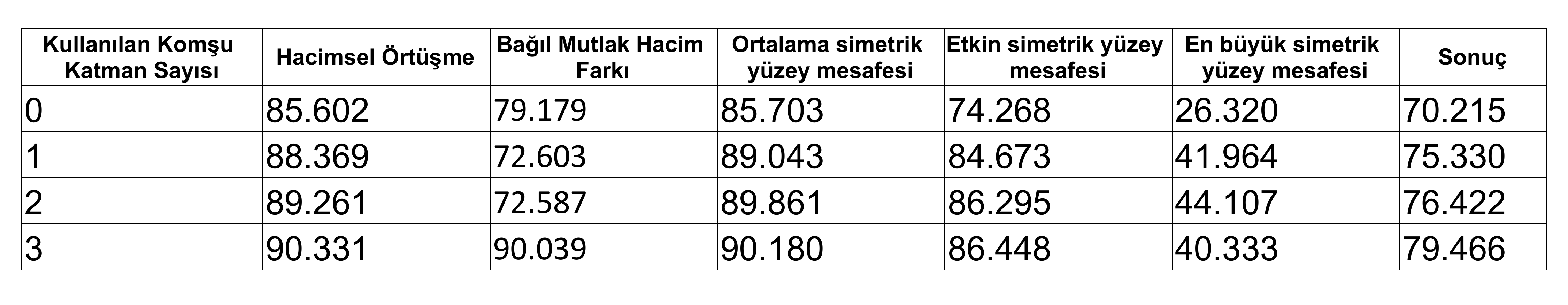}}
\end{table} 

Tablo \ref{Table:postProcEffect}'de son işleme olarak 2B ayrık giderme uygulandığında elde edilen sonuçlar görülmektedir. Tablo \ref{Table:NeighborSlicesEffect}'de verilen son işlemesiz halleri ile karşılaştırıldı-ğında sonuç puanlarının yaklaşık 20 puan arttığı görülmektedir. Bu da, sunulan yöntemin aslında basit bir işlem adımıyla çok daha başarılı sonuçlar verebileceğini göstermektedir. Ayrıca, komuşulukların kullanılmasının performansa olan olumlu etkisi benzer şekilde bu tabloda da gözlenmektedir. 

Son olarak, 2B yerine 3B ayrık giderme son adımı uygulandığında sonuçlara olan etkisi gösterilmiştir. Tablo \ref{Table:siu2018LiverChallenge}'de önerilen yöntemin (METU MM LAB v5) SIU Karaciğer Bölütleme Algoritmaları Yarışması'nda sunulan diğer yöntemlerle karşılaştırmalı sonuçları verilmektedir. Burada, önerilen yöntem sadece verilen kısıtlı veri seti ile eğitilmiştir. İTÜ Vision Lab-v10 algoritmasında ise farklı veri setleri ile eğitim veri seti daha geniş tutulmuştur. Bu da, önerilen yöntemin limitli eğitim seti ile dahi en iyi başarımı verebileceği ve MR görüntü bölütleme problemi için en iyi yöntem olabileceğinin göstergesidir. 

\begin{table}[!h]
	\centering
	\caption{Yarışmada sunulan en başarılı yöntemlerin sonuçları.} \label{Table:siu2018LiverChallenge}
	\vspace{-10pt}
	{\shorthandoff=%
		\includegraphics[width=11cm]{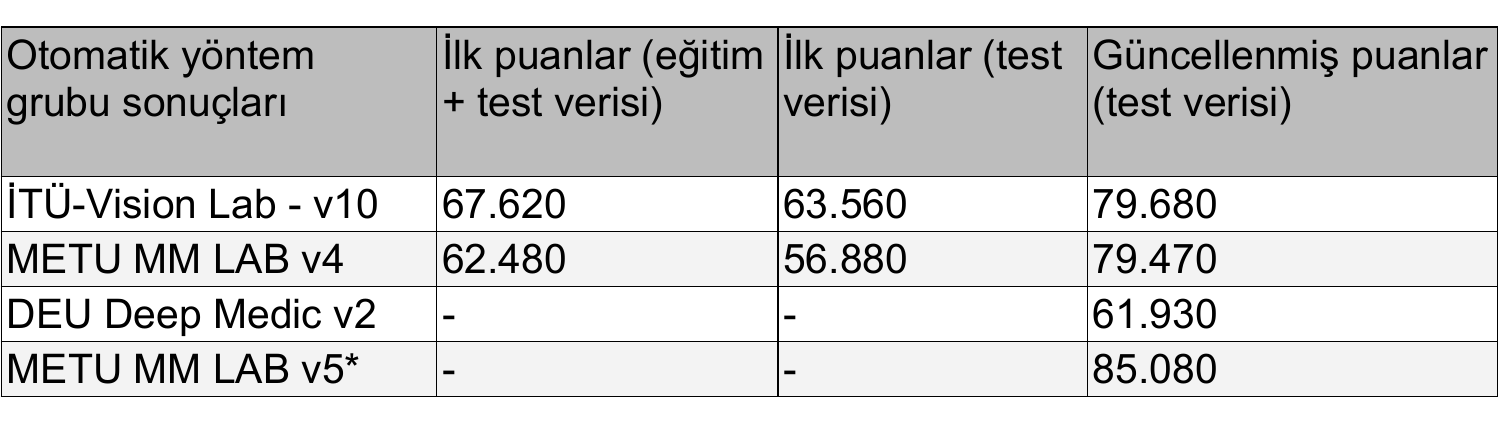}}

\end{table} 

\blfootnote{*2 boyutlu ayrık giderme 3 boyutlu olarak uygulandığında elde edilen sonuç}

\section{Yorumlar}

Tıbbi görüntüler 3 boyutlu olarak toplandığından ve değerlendirme sırasında bu bilgiden faydalanılması gerektiğinden, 3 boyutlu otomatik yöntemlerin başarımı artırabildiği gösterilmiştir. Ayrıca, önerilen yöntemde, Koşullu Çekişmeli Ağlar eğitime dahil edilmiştir.  Bu sayede doku sınırlarında, ağ bir koşullama kurarak, yöntemin eğitimi geliştirilmiş ve başarımı artırılmıştır. Son olarak, elde edilen bölütleme görüntülerinde ayrıkların temizlenmesinin başarımı arttırdığı özetlenmekdir. Önerilen yöntemin testleri sırasında sadece karaciğer bölütleme problemi kullanılsa da, bu yöntem herhangi bir MR görüntü bölütleme problemine rahatlıkla uygulanabilir.

\section{Teşekkürler}

Önerilen modelin eğitimi ve testleri sırasında kullanılan hibe edilmiş GPU kartları için yazarlar NVIDIA şirketine memnuniyetlerini sunar. Ayrıca, yarışma sırasında ve sonrasında değerli fikir alışverişlerinden dolayı Sn. Alper Selver'e ve Sn. Emre Kavur'a teşekkür eder.

\bibliographystyle{ieeetr} 
\bibliography{samplepaper} 
\end{document}